**Article**

# Neural Evidence and Behavioural Welfare Economics: When Can Neuroeconomics Inform Policy?

Louis Yiven Zhu[1]

[1] Student (BSc Science and Technology Studies, 2023-2026), Department of Science and Technology Studies, University College London, United Kingdom; yiven.zhu@oii.ox.ac.uk; ORCID (0009-0001-5579-0340)

**Abstract**
Neuroeconomics appears to offer welfare economics an evidential supplement, since neural data may help distinguish choices that express genuine preferences from choices distorted by addiction, framing or self-control failure. This paper asks when such evidence can legitimately inform welfare judgements for policy. It develops a conceptual framework linking three levels of analysis, namely neural signals, computational decision models and normative welfare criteria. The central argument is that neural data can constrain welfare analysis only under four conditions, covering the validity of the neural measure, the identifiability of the decision model, the explicit statement of the welfare criterion, and the institutional context of any policy use. Applying this framework to addiction, neuromarketing and environmental policy, and working through the United Kingdom's Tobacco and Vapes Act 2026, a major intervention justified without any neural measurement, the paper develops a Neuroeconomic Welfare Inference Checklist for neural evidence. Neural data can improve behavioural welfare economics by clarifying the mechanisms behind choice, and they cannot replace normative argument about what counts as welfare. The same reasoning applies to artificial agents trained on human feedback, whose internal reward signals are computational quantities and not welfare measures.

# 1. Introduction

Neuroeconomics is the study of the neurobiological and computational basis of value-based decision-making (Rangel, Camerer and Montague, 2008; Camerer, Loewenstein and Prelec, 2005). Early work on dopaminergic reward-prediction errors and cortical value representations suggested new ways to think about intertemporal choice, risk and self-control (Schultz, Dayan and Montague, 1997).

Welfare economics and public policy have begun to draw on these findings. Behavioural public economics asks how to do welfare analysis when preferences are inconsistent or context-dependent, and neuroeconomics appears to offer mechanistic evidence about which choices reflect 'true' interests and which reflect mistakes or cue-triggered lapses (Bernheim and Rangel, 2004, 2007; Bernheim, 2009). Applications now range from consumer protection and retirement saving to health and environmental policy, alongside a commercial ecosystem in neuromarketing (Hsu and Yoon, 2015; Alsharif and Mohd Isa, 2024).

This expansion raises a central problem. Neural data are noisy, paradigm-dependent and theory-laden, whereas welfare is a normative construct encoding judgements about when a person is better or worse off (Poldrack, 2011; Krajbich and Dean, 2015). That a choice correlates with activation in the ventral striatum or in ventromedial prefrontal cortex (vmPFC) does not tell us whether it advances the person's welfare, and without a framework the authority of brain images risks re-labelling contested welfare judgements as scientific facts (Gul and Pesendorfer, 2008; Bernheim, 2009).

We argue that neural data can inform welfare only via a clearly specified computational model and a clearly defended normative standard. Brains implement algorithms, and neuroeconomics allows us to infer aspects of those algorithms from neural data and behaviour. Welfare analysis then requires an additional step, which is to decide which outputs count as welfare-relevant in which contexts.

We make three contributions. First, we set out a framework in which the relation between neural signals, computational decision variables and welfare criteria is one-directional and conditional, so a welfare claim can be traced back through the model that licenses it to the measurement that supports it. Second, we state the conditions under which neural evidence can constrain a welfare judgement and sets them out as six steps, extending to the welfare question the model-validation standards that computational modelling and neuroimaging have developed for the positive question. Third, we show that the framework reaches beyond human choice, since artificial agents trained by reinforcement learning also carry internal reward and value signals, and the same reasoning shows why those cannot stand in for a welfare criterion.

We distinguish three levels, namely neural data such as spikes and the blood-oxygen-level-dependent (BOLD) signal, computational models of values, prediction errors and decision rules, and welfare judgements about what is better or worse for the person. Neural evidence can constrain welfare analysis only by way of a computational model, and welfare claims always require an additional normative step (Bernheim, 2009; Bernheim and Taubinsky, 2018). Sections 3 and 4 make this concrete using an actor-critic architecture and its neural implementation (Figures 1 to 3). Appendix A collects the formal model and Appendix C lists every symbol.

In plain terms, brains compute when people choose, and some of those computations leave measurable traces. A brain measurement can therefore be read as evidence about a quantity inside a model of how someone decides, and that is useful for welfare economics because the models that treat certain choices as mistakes, and so justify intervening, are models with quantities of this kind. What a brain measurement cannot do is settle which of a person's choices count for their welfare, since that question is answered by argument about what makes someone better off. Our contribution is to identify the steps in that chain that neural evidence can support and which it cannot.

Figure 4 sets out the same argument as a diagram, with the three levels as bands and the four conditions of Section 4 as gates that any inference from neural data to a welfare claim must pass.

# 2. Neuroeconomics, Welfare and NeuroAI

## 2.1. Neuroeconomics as value-based decision science

Rangel, Camerer and Montague (2008) decompose value-based decisions into five processes, which are representing the problem, valuing options, selecting an action, evaluating the outcome and updating value representations.

Empirically, orbitofrontal cortex and vmPFC encode subjective values of options, often in a 'common currency' across reward modalities, while ventral striatum and midbrain dopamine neurons show activity consistent with reward-prediction errors (Schultz, Dayan and Montague, 1997; Rangel, Camerer and Montague, 2008).

A key implication is that these neural variables track subjective evaluations, contingent on expectations, reference points and affective states (Camerer, Loewenstein and Prelec, 2005). High striatal activation is not a direct readout of welfare, since it reflects context-dependent value signals.

## 2.2. Welfare economics and behavioural public economics

Standard welfare economics identifies welfare with the satisfaction of coherent, stable preferences revealed in choice. When preferences are complete, transitive and context-independent, a utility function represents them and can serve as a welfare criterion.

Behavioural economics documents systematic violations, among them present bias, framing effects, loss aversion and self-control problems, and behavioural public economics develops welfare notions that accommodate them (Bernheim and Rangel, 2007, 2009; Bernheim and Taubinsky, 2018).

One influential strand distinguishes welfare-relevant from mistake-driven choices. For example, an 'as-judged-by-themselves' criterion seeks to recover the preferences individuals would endorse under better conditions, free of errors and self-control failures (Thaler and Sunstein, 2008; Bernheim and Rangel, 2009).

Neuroeconomics appears to offer mechanistic evidence about which choices express underlying goals. Addiction is a central case (Bernheim and Rangel, 2004; Becker and Murphy, 1988). Whether such evidence can discriminate between competing accounts is an identification question, which Appendix B answers.

## 2.3. NeuroAI and value-learning systems

Reinforcement learning offers a natural bridge between neuroeconomics and artificial intelligence. Temporal-difference learning updates value estimates using prediction errors, a formalism closely matching dopamine neuron activity in conditioning and choice tasks (Sutton and Barto, 2018; Montague, Dayan and Sejnowski, 1996).

Dabney et al. (2020) show that dopamine populations in mice encode a distributional value signal, with different neurons tuned to different reward quantiles, mirroring distributional reinforcement-learning algorithms in machine learning. Sadeh and Clopath (2025) describe this two-way traffic as the core of NeuroAI, in which artificial-intelligence tools accelerate neuroscience while neural principles inform new architectures.

Conceptually, both brains and artificial agents can be treated as value-learning systems that maintain value functions, decision rules and learning rules, and welfare questions arise when we ask which values they ought to learn. Section 6 takes this up as an implication of the framework.

# 3. A Formal Mapping from Neural Signals to Welfare Claims

This section sets out a compact framework that makes explicit the steps from neural data to welfare-relevant judgements. Its aim is to show where normative assumptions enter. Appendix A gives the model in full.

## 3.1. Decision and learning model

Consider a sequential decision problem with discrete time $t$. The individual occupies states $s_t \in S$, chooses actions $a_t \in A$, receives reward $r_t$ on the transition out of $s_t$ and moves to $s_{t+1}$. The critic holds a value function $V_\theta(s)$ that estimates the value of the decision rule the individual actually implements,

$$V^{\pi_\phi}(s) \;=\; \mathbb{E}\big[\textstyle\sum_{k=0}^{\infty} \gamma^k\, r_{t+k} \;\big|\; s_t = s,\, \pi_\phi\big], \qquad V_\theta(s) \approx V^{\pi_\phi}(s).$$

In words, the critic tracks how good a situation is, judged by the rewards the person actually goes on to collect while behaving as they actually behave and not as they ideally would.

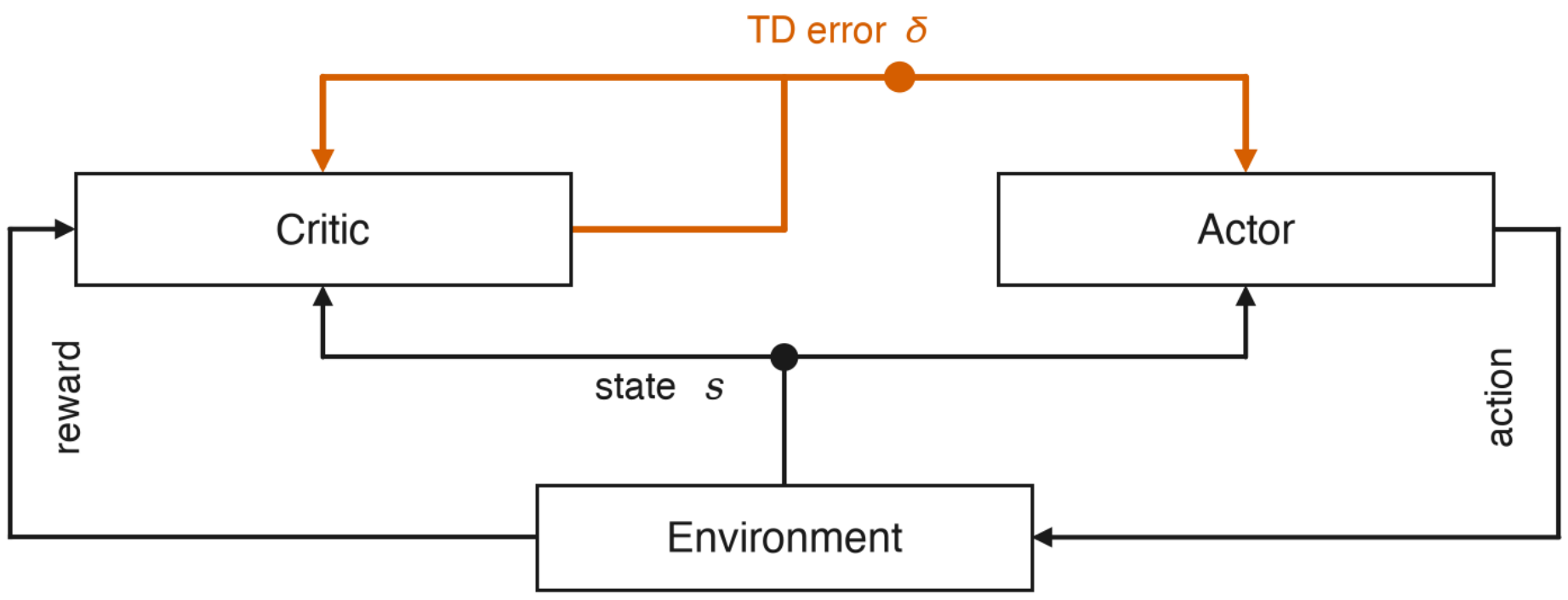

Figure 1: The actor-critic architecture. The critic estimates the value of the rule the individual implements, equation (1); the actor selects actions by the softmax rule, equation (2). The temporal-difference (TD) error of equation (3), shown in orange, is computed by the critic and updates both the critic, equation (9), and the actor, equation (10). Drawn by the author after Sutton and Barto (2018) and Barto, Sutton and Anderson (1983).

Actions follow a softmax decision rule over the actor's own action-preference function $h_\phi$,

$$\pi_\phi(a \mid s_t) \;=\; \frac{\exp\{\beta\, h_\phi(s_t, a)\}}{\sum_{a' \in A} \exp\{\beta\, h_\phi(s_t, a')\}},$$

where $\beta$ is an inverse-temperature parameter controlling sensitivity to preference differences. In words, actions the person prefers are chosen more often, and the parameter governs how sharply preference translates into choice.

Learning is driven by the temporal-difference error

$$\delta_t \;=\; r_t + \gamma\, V_\theta(s_{t+1}) - V_\theta(s_t),$$

where $\gamma$ is the individual's discount factor. The critic raises its estimate of the value of the state just left when the outcome is better than expected and lowers it when the outcome is worse, at a rate set by its learning rate (Appendix A, equation (9)). The actor makes the action just taken more likely when the same error is positive (Appendix A, equation (10)). In words, a better outcome than expected makes the action just taken more likely next time. This is the architecture of Figure 1 (Sutton, 1988; Barto, Sutton and Anderson, 1983), and it is implemented in artificial agents and in biologically plausible network models alike (Sutton and Barto, 2018; Potjans, Diesmann and Morrison, 2011).

## 3.2. Neural signals as noisy encodings of computational quantities

Let $n_t$ denote neural measurements at time $t$, for example firing rates or BOLD activity. In the simplest case these are noisy encodings of latent computational variables, the dopaminergic channel encoding the prediction error,

$$n_t^{\mathrm{DA}} \;=\; f(\delta_t) + \varepsilon_t,$$

and the prefrontal measurement stands in the same relation to the value signal (Appendix A, equation (12)). Here $f$ and $g$ are link functions drawn from a class fixed in advance and $\varepsilon_t, \eta_t$ are noise terms. Appendix B.2 states what the choice of class buys.

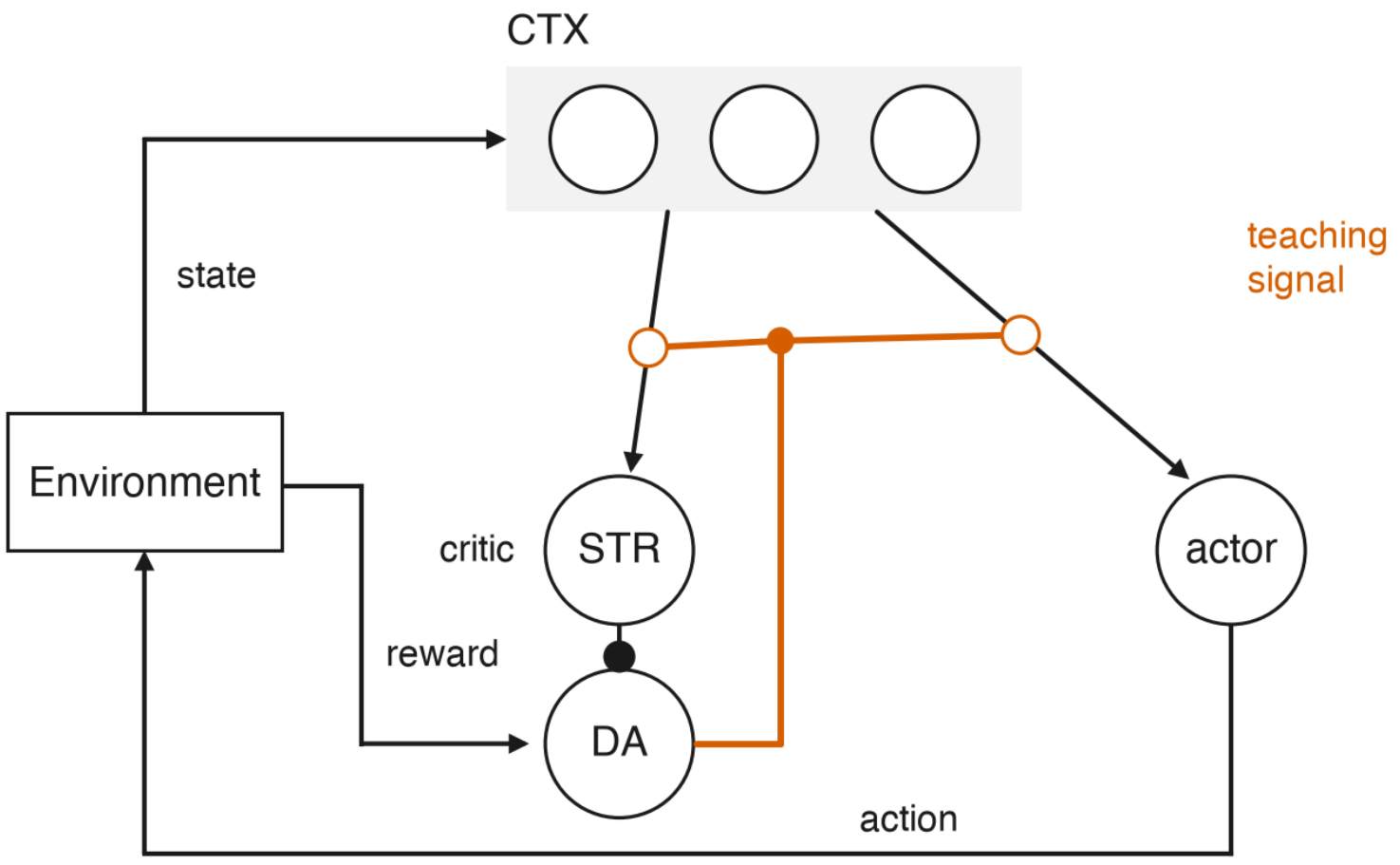

Figure 2: A neuronal actor-critic circuit. Cortical populations (CTX) carry state information to a striatal critic (STR) and to an actor population. The striatal prediction inhibits dopaminergic neurons (DA), which receive reward. Arrowheads are excitatory, the filled dot inhibitory, the open circles modulatory. The orange teaching signal branches from DA onto both cortical projections. Adapted from Potjans, Diesmann and Morrison (2011, Fig. 2), CC BY 4.0; redrawn by the author.

Figure 2 extends the abstract actor-critic of Figure 1 to a neuronal implementation (Potjans, Diesmann and Morrison, 2011). Cortical populations encode state information, the critic is implemented in striatal and dopaminergic neurons, and the actor is a pool of action-selective neurons. Cortical inputs drive the striatal and dopaminergic populations, and dopaminergic neurons broadcast a neuromodulatory signal that globally modulates plasticity at cortico-striatal and cortico-actor synapses. The dopaminergic signal therefore plays the role of a TD-error-like teaching signal in the actor-critic loop.

Empirically, phasic dopamine activity does resemble a TD prediction error in classic conditioning tasks. Before learning, unpredicted rewards elicit brief dopaminergic bursts; after learning, the burst shifts to the predictive cue and disappears at reward time; omission of an expected reward produces a dip below baseline (Schultz, Dayan and Montague, 1997).

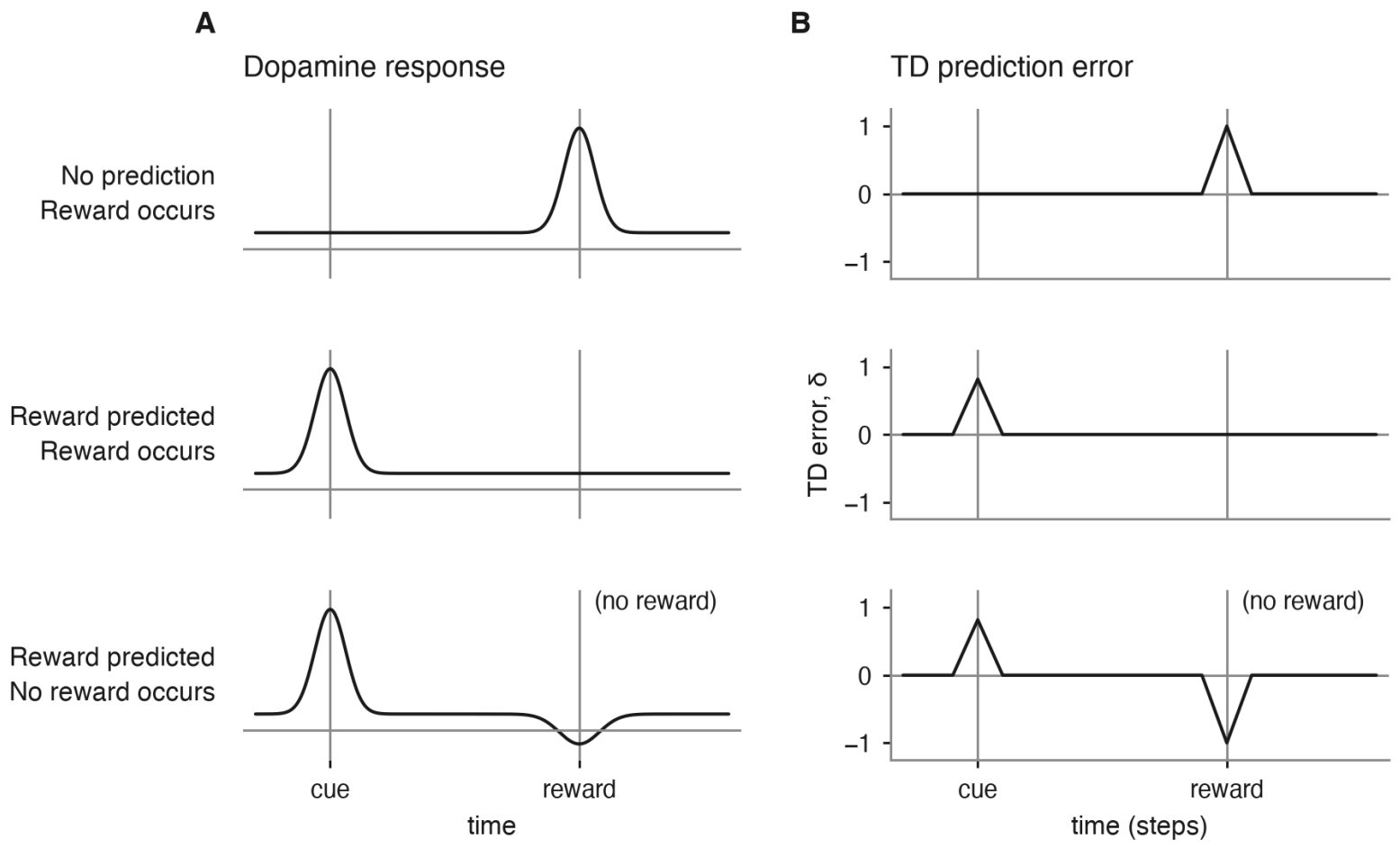

Figure 3: Dopamine responses and temporal-difference prediction errors. Panel A, schematic dopaminergic firing in the three canonical conditions, after Schultz, Dayan and Montague (1997), in arbitrary units with no scale. Panel B, prediction errors computed by the author from the model of Section 3.1, with a discount factor of 0.98 and a cue-reward delay of ten steps. Reward timing follows equation (3). Parameters and code are in Appendix E.

The qualitative match in Figure 3 justifies, at least in this paradigm class, modelling dopaminergic activity as an encoding of $\delta_t$ (Lloyd and Dayan, 2015).

Neural signals therefore provide noisy constraints on the latent variables of a reinforcement-learning model. Dopamine gives evidence about $\delta_t$, vmPFC and ventral striatum about $V_\theta$, and prefrontal regions about the control signals that govern $\pi_\phi$. Whether and how those constrained models inform welfare analysis is the subject of Section 4.

### 3.3. Welfare criteria and true interests

Suppose we want to evaluate whether a policy $p$ that changes the decision environment makes individuals better off. Behavioural-welfare frameworks distinguish the implemented decision process from a welfare criterion (Bernheim and Rangel, 2009; Bernheim and Taubinsky, 2018).

Let $U(s,a)$ denote a welfare-relevant utility function, and let $V_\theta$ and $\pi_\phi$ describe the value function and decision rule the person actually implements. These need not coincide, and in particular there may be states in which $\pi_\phi$ selects actions that reduce $U$.

A dual-self self-control model makes this explicit. With $u^L$ a long-run utility and $u^S$ a short-run utility, implemented instant utility is

$$U^{\text{impl}}(s,a) \;=\; \lambda(s)\,u^L(s,a) + (1-\lambda(s))u^S(s,a), \qquad \lambda: S \to [0,1],$$

where $\lambda(s)$ reflects the influence of long-run considerations. In addiction, cue exposure may transiently reduce $\lambda$, shifting weight towards the short-run self (Bernheim and Rangel, 2004).

The reward that drives learning is the implemented instant utility evaluated at the realised state and action,

$$r_t \;=\; U^{\text{impl}}(s_t, a_t).$$

Equation (6) is the only link the argument needs between the learning model of Section 3.1 and the welfare vocabulary of this section, and it is a modelling assumption and not a normative one. What the person learns from is the utility they act on.

A welfare criterion may take $u^L$ as normative and treat episodes dominated by $u^S$ as mistakes. Policies are scored and compared by

$$W(p;\,U) \;=\; \mathbb{E}^p\left[\sum_{t=0}^{T} \tilde{\gamma}^{\,t}\; U(s_t, a_t)\right],$$

$$\Delta(p, p_0;\, U) \;=\; W(p;\,U) - W(p_0;\,U).$$

In words, an intervention is scored by the discounted welfare-relevant utility the person accumulates while behaving as they actually behave under it, and two interventions are compared on that score. The discount factor $\tilde{\gamma}$ and the horizon $T$ belong to whoever is doing the evaluation, and they are distinct from the agent's own discount factor $\gamma$ in equation (3), which the framework treats as a behavioural parameter that cue exposure and self-control failure may themselves distort. Welfare is defined in terms of $U$, and not in terms of $V_\theta$ or $n_t$. Neural and behavioural data tell us about $V_\theta$, $\pi_\phi$ and the environments in which they are distorted, and they do not determine $U$.

## 4. When Do Neural Data Legitimately Inform Welfare?

The framework of Section 3 invites the question an economist asks first. Choice data, surveys and field experiments already establish that many smokers hold inconsistent intertemporal preferences and that most report wanting to stop, so it is fair to ask what a brain measurement adds. The answer has three parts and each is narrower than the enthusiasm around neuroeconomics suggests. Neural measurements enter the likelihood as further observables, so two decision models that fit the same choices can be told apart when they imply different signals for those choices. They can locate the states in which implemented choice and welfare come apart, since cue reactivity is a property of a moment that self-report recovers poorly once the moment has passed. And a mechanism validated at the level of computational variables gives some warrant for extrapolating to policy environments the data do not contain. There are two circumstances, both common, in which neural data add nothing at all. Where a behavioural sufficient statistic already pins down the welfare-relevant bias parameter, as it does for an optimal corrective tax, no additional moment changes the policy conclusion (Allcott, Lockwood and Taubinsky, 2019). And no measurement of any kind selects the welfare criterion itself.

Within this framework, neural data can constrain welfare analysis only under four conditions.

## 4.1. Measurement and representational validity

The actor-critic architecture in Figures 1 and 2, together with the empirical mapping between dopamine and TD error in Figure 3, motivates treating neural data as evidence about $\delta_t$, $V_\theta$ and $\pi_\phi$. For such evidence to bear on welfare, three further requirements apply.

Task-level validation asks for convergent evidence that $n_t$ tracks $\delta_t$ or $V_\theta$ across tasks of the relevant class, and not in a single paradigm (Schultz, Dayan and Montague, 1997; Montague, Dayan and Sejnowski, 1996).

Representational specificity asks that the encoding in equation (4) be specific enough that alternative computational models do not explain the data equally well, since otherwise neural data favour no particular welfare analysis (Poldrack, 2011; Krajbich and Dean, 2015). The strength of the identification result below depends on what is assumed about the encoding functions, and Appendix B sets out the two cases, affine maps giving cardinal identification and monotone maps giving only ordinal identification (Caplin and Dean, 2008).

Robustness asks that encoding be stable across individuals and contexts, or that systematic variation be modelled. Without these three, neural data cannot be treated as measures of value or prediction error.

## 4.2. Model identifiability

Even with a plausible encoding, there is often a model identifiability problem, since multiple decision and learning models can reproduce the same pattern of behaviour and neural responses (Wilson and Collins, 2019; Turner et al., 2013). Two structures are observationally equivalent on choice when they induce the same distribution over action paths (Appendix B, equation (16)).

Neural measurements can break this equivalence. Proposition 1 in Appendix B.4 shows that two structures indistinguishable on choices separate on the joint likelihood of actions and neural measurements exactly when they imply different distributions of the encoded computational variables given those choices, and Remark 1 there shows that the welfare criterion itself is identified by no data of either kind.

A second condition follows. The model class must be sufficiently constrained in the domain of application, and alternative models with distinct welfare implications must be either empirically ruled out or explicitly acknowledged (Bernheim and Rangel, 2009; Bernheim, 2009). Appendix D records which equations and assumptions each step relies on.

## 4.3. Normative transparency

A third condition concerns the normative move from computational variables to welfare. When authors claim that one neural pattern reflects a ‘true’ valuation and another a ‘biased’ one, they are selecting a welfare function $U$ and classifying some states as mistake states, and that judgement is a separate step from the neuroscientific mapping of states to mechanisms (Bernheim and Rangel, 2007; Bernheim and Taubinsky, 2018).

Any paper or policy drawing welfare conclusions from neural data should therefore state the criterion explicitly, justify it, and explain how the data identify contexts in which welfare and implemented choice diverge.

### 4.4. Policy relevance and institutional context

Translating individual-level neural evidence into policy involves further steps. Policies operate at scale, under constraints of fairness, privacy and feasibility, and using neural data to justify targeted mandates raises distributional and political concerns (Bernheim and Rangel, 2007; Bernheim and Taubinsky, 2018).

Any proposed use should specify how the evidence would be operationalised, whether classification on neural traits is legitimate, and how uncertainty and heterogeneity will be handled.

Table 1 operationalises these four conditions as six steps, splitting model identifiability into the specification of the model (Step 2) and its assessment against rival models (Step 4), and normative transparency into the choice of criterion (Step 1) and the location of divergences (Step 5).

## 5. Case Domains

This section applies the framework to three domains in which neural data have been proposed as inputs to welfare analysis. The first is worked through all six steps against an enacted policy.

### 5.1. Addiction and self-control

Drug exposure alters dopaminergic signalling and cue reactivity, drug-related cues acquire motivational salience, and many studies report reduced prefrontal control in addicted individuals (Everitt and Robbins, 2016; Redish, 2004). The Tobacco and Vapes Act 2026 is worked here because it is an enacted intervention resting on that picture, and Appendix F gives the provision, paragraph and page behind each step below.

**Step one, the welfare criterion.** The Department of Health and Social Care (DHSC) impact assessment adopts a health-outcome standard. It values avoided mortality in quality-adjusted life years at £70,000, discounts health gains at 1.5 per cent and other monetised effects at 3.5 per cent, and appraises 30 years from implementation on 1 January 2027, projecting 154,800 deaths and 472,950 cases across four smoking-related diseases avoided in England by 2100 (DHSC, 2024). The command paper defends that standard in language this framework recognises, noting that over 80 per cent of smokers started before the age of 20 and ‘have had their choices taken away by addiction’ (DHSC, 2023, p. 2). Adults who may lawfully be sold tobacco today keep that entitlement, so the design privileges the long-run self of a future cohort and leaves current smokers’ revealed preferences intact.

**Step two, the computational model.** In the notation of Section 3 and Appendix A, adolescent initiation is a regime in which cue-associated value $\kappa C(s, a)$ grows and the weight $\lambda$ on the long-run self falls, so implemented choice drifts away from $u^L$. The rival specification is rational addiction (Becker and Murphy, 1988), under which the same consumption path maximises a stable intertemporal objective and no mistake is committed.

**Step three, neural encodings.** No neural measurement enters the appraisal. One neuroscience review is cited, at paragraph 11 footnote 22 (Leslie, 2020), and it does qualitative work in the rationale for intervention. Meta-analyses of functional magnetic resonance imaging (fMRI) studies report reliable cue-elicited responses in ventral striatum, amygdala and medial prefrontal cortex among smokers (Engelmann et al., 2012; Lin et al., 2020), and reviews of adolescent nicotine exposure describe altered dopaminergic and cholinergic maturation (Yuan et al., 2015). That evidence comes from cue-viewing paradigms and animal models, so it supports a cue regime without identifying $\delta_t$ or $V_\theta$ inside the purchase decisions the statute governs, and condition one is met in a weak form only.

**Step four, identifiability.** Behavioural evidence alone leaves the two accounts of adolescent uptake observationally equivalent. Gruber and Kőszegi (2001) show that smokers are forward-looking while time-inconsistent preferences fit the same behaviour. A rational-addiction consumer and a cue-triggered decision-maker generate the same uptake path, the same response to anticipated price changes and the same expressed wish to stop. The two accounts do imply different distributions of the prediction-error signal for those same choices, so on the joint likelihood of actions and neural measurements they separate, provided both are estimated on a common environment and the encoding maps are drawn from one class fixed in advance (Appendix B, Proposition 1). Neither condition holds across the adolescent samples the cue-reactivity literature reports, so the separation is available in principle and not on the present evidence.

**Step five, welfare-relevant divergences.** Divergence between implemented choice and $U$ concentrates in adolescence, where initiation happens and where the assessment records that dependence forms fastest (DHSC, 2024). A rule keyed to date of birth reaches that margin, restricting the cohort whose choices the model classifies as distorted and leaving the rest alone.

**Step six, implementation.** Section 1 of the Act classifies by date of birth and not by any measured trait, which is what the policy-context condition asks of a legitimate classification. Two adults born days apart hold different entitlements; licensing and age verification place enforcement on retailers (ss. 16 to 22); and the review clause promises only evidence on economic and health impacts within four to seven years (s. 165). New Zealand legislated the same design in December 2022, for the same birth cohort and with the sales prohibition due to bite in 2027, and repealed it in March 2024 before the provisions took effect (House of Lords Library, 2024), which is a reminder that the welfare case for a generational ban has to survive a change of government as well as a checklist. Neural evidence would therefore strengthen two steps of this appraisal and could not reach the others. It would strengthen Step three, by validating that the measured signal encodes the computational variable the model names, and Step four, by separating the two accounts of uptake on the joint likelihood. It cannot bear on Step one, which fixes the welfare criterion by argument, or on Step six, where the legitimacy of classifying by date of birth is a question about institutions.

## 5.2. Neuromarketing and behavioural public policy

Neuroeconomics and behavioural insights now inform public policy and commercial practice alike, including neuromarketing and personalised advertising (Hsu and Yoon, 2015). Public bodies use fMRI and electroencephalography (EEG) to find framings that engage self-control networks, and firms use the same methods to maximise reward-related activation.

The same methods serve opposite ends. Recommendation algorithms that maximise engagement reshape the reward landscape in which human learning systems operate, and over time they may lead to over-consumption of immediately rewarding but harmful content (Kleinberg, Mullainathan and Raghavan, 2023). Platforms choose the reward function and the state-to-cue mapping while human value functions adapt.

Australia's Online Safety Amendment (Social Media Minimum Age) Act 2024 is the closest enacted analogue to the tobacco case, and it too was justified without neural evidence. Neural data would support welfare claims here only alongside an explicit criterion at Step 1, a model of how design shifts behaviour away from it at Step 2, and transparency at Step 5.

## 5.3. Environmental and energy policy

Environmental and energy policy is a further domain (Sawe and Chawla, 2021). Neuroimaging studies examine responses to framings of climate risk, and strong activation in regions linked to affective value is sometimes read as latent concern that choices under-express.

The framework suggests caution here since neural data can show that certain messages more strongly engage valuation or moral-concern circuits, and can reveal heterogeneity across individuals (Sawe and Chawla, 2021). Moving from that to a conclusion about how policy should weight environmental quality requires a normative step. The checklist applies at Steps 3, 4 and 5.

# 6. A Neuroeconomic Welfare Inference Checklist and Implications for NeuroAI

Drawing the argument together, Table 1 summarises a practical checklist for using neural data in welfare analysis. Figure 4 shows the three levels and the gates an inference has to pass, and Table 1 with its caption give the mapping from steps to gates, so the checklist can be read as an operational form of the conditions of Section 4.

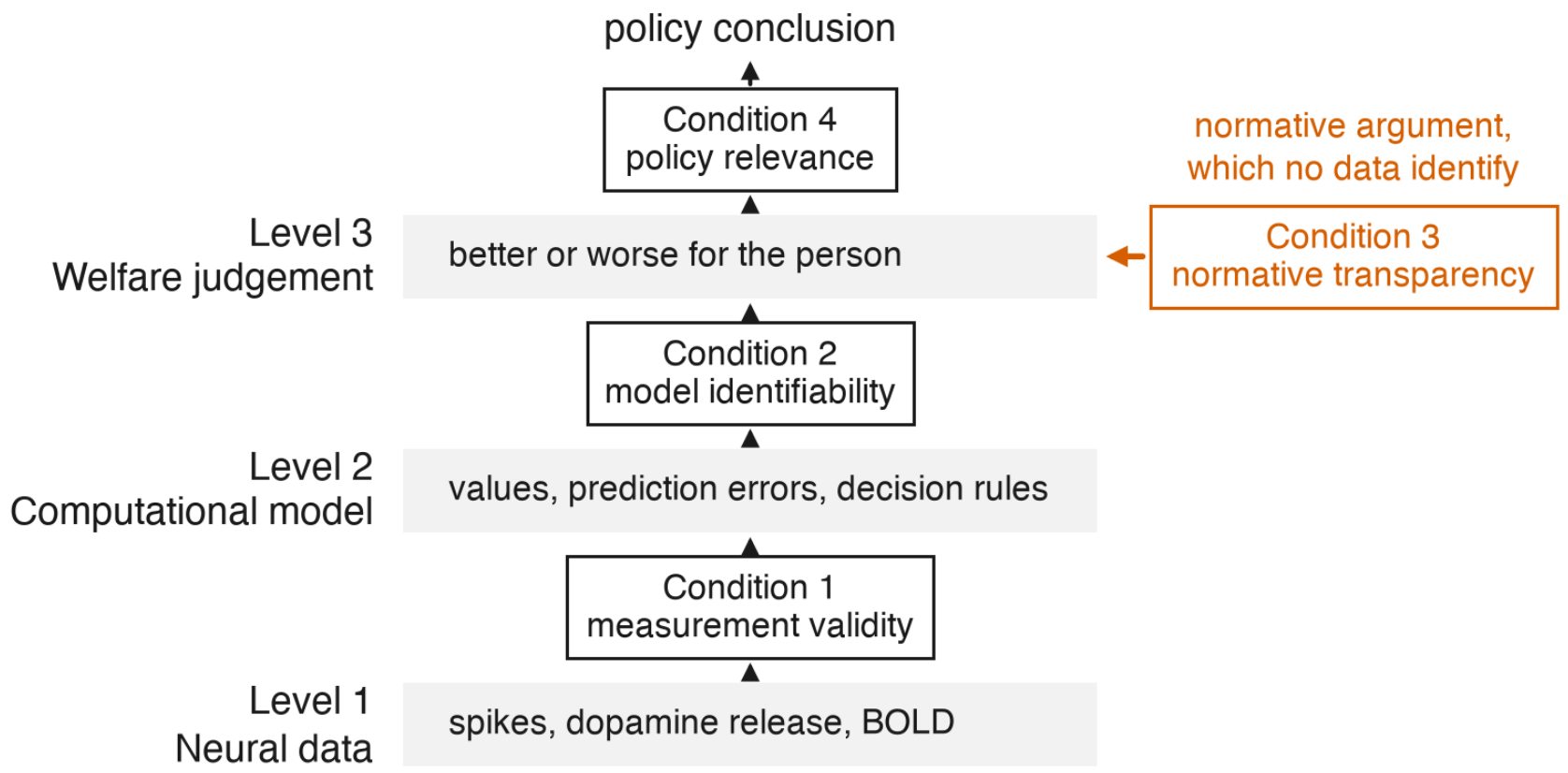

Figure 4: When neural evidence can constrain a welfare judgement. Three levels as bands. Condition 1 gates neural data into the model, Condition 2 the model into a welfare judgement, Condition 3 gates the normative input, shown in orange, which no datum identifies, and Condition 4 the output into a policy conclusion. Table 1 maps the six steps to the conditions. All four must hold. Original figure by the author.

Table 1: Neuroeconomic Welfare Inference Checklist. The Condition column gives the condition of Section 4 that each step operationalises. Appendix D gives the full mapping from steps to equations and assumptions.

| **Step** | **Condition (Section 4)** | **Details** |
|---|---|---|
| 1. Define the welfare criterion $U$ | Normative transparency (4.3) | State whether welfare is long-run preferences, experienced utility or another standard. Equations (7) and (8). |
| 2. Specify the computational model | Model identifiability (4.2) | State the model and its parameters $\theta, \phi$. Equations (1) to (3), (5) and (6). |
| 3. Validate neural encodings | Measurement validity (4.1) | Give evidence that neural variables encode quantities such as $\delta_t$ and $V_\theta$. Equation (4). |
| 4. Assess model identifiability | Model identifiability (4.2) | Consider rival models fitting the same data, and whether neural evidence separates them. Appendix B, Proposition 1. |
| 5. Locate welfare-relevant divergences | Normative transparency (4.3) | Identify states where implemented choice diverges from $U$, using neural evidence on mechanisms only. Equation (5), and Appendix A, equation (15). |
| 6. Analyse policy implementation | Policy relevance (4.4) | Explain how neural classifications would be implemented, addressing fairness, privacy and legal constraints. |

For NeuroAI the checklist has three implications.

First, designers of systems that assist human decision-makers can adopt an explicit welfare criterion $U$ and treat human neural and behavioural data as evidence about $V_\theta$ and $\pi_\phi$, with the checklist disciplining how that evidence is used.

Second, optimising neural activation is not optimising welfare. A system that maximises expected striatal activation might push users towards behaviours with high short-run reward and low long-run welfare, so welfare-aligned design should optimise a defensible $U$, with neural data serving as one class of inputs.

Third, the same reasoning applies to systems trained on human feedback. Reinforcement-learning agents are commonly trained against a reward model fitted to human preference judgements (Christiano et al., 2017), and platforms that optimise engagement choose the reward signal under which human value functions then adapt (Kleinberg, Mullainathan and Raghavan, 2023). In both cases an internal reward signal stands in for what is good for the user, and the framework explains why that substitution needs an argument. Such a signal is a computational quantity in an artificial system exactly as striatal activation is one in a human system. A designer who wants the system to serve a user's interests must specify and defend a welfare criterion separately, and the six steps govern how evidence about that user may be used to adapt it.

Three limitations bear on how far this framework can be taken. The paper is conceptual, and although the checklist is written to be applied, it has not been used to audit a body of published work, so its discriminating power is untested. The dopamine and prediction-error correspondence on which the measurement condition leans is strongest in the conditioning paradigms in which it was established, and its extension to the richer choices that policy concerns is an assumption. The checklist also presumes a class of decision and learning models that applied welfare analysis has not standardised, so two analysts applying it to the same policy may specify Step 2 differently and reach different conclusions at Step 4.

# 7. Conclusion

Neuroeconomics is sometimes presented as if it could directly deliver welfare-relevant quantities by reading the brain. This paper argues that such expectations are misplaced. Neural data are valuable, and they inform welfare only through computational and normative lenses.

By formalising the path from neural signals, through reinforcement-learning and dual-self models, to welfare criteria, I have derived conditions under which neural data can legitimately inform policy. These emphasise representational validity, model identification, normative transparency and institutional context. The Tobacco and Vapes Act 2026 shows a major intervention justified without any neural measurement, and the checklist identifies the two steps at which such measurement could have contributed.

The alignment with NeuroAI is natural, since reinforcement learning, prediction-error coding and distributional value representation provide a shared language for brains and machines (Doya, 2007; Dabney et al., 2020). Welfare remains a separate layer that requires philosophical and economic argument. The measurement problem of Section 4.1 has no artificial analogue, and the normative problem of Remark 1 transfers in full.

Neural data can refine our models of how people choose and illuminate the mechanisms behind departures from rationality, and embedded in transparent welfare frameworks such models can support more precise and humane policy design. What neural data cannot do is relieve us of the need to argue about what counts as a good life, and about whose judgement should prevail when brain signals and choices diverge (Bernheim, 2009; Krajbich and Dean, 2015).

**Notes**

Competing interests: The author declares no competing interests.
Data and code availability: No new empirical data were generated for this article. Panel B of Figure 3 is computed by the author, and the code that produces it is released as fig3_td_simulation.py, together with its output file fig3_td_simulation_output.csv and the script that renders the figures, at https://github.com/louisyzhu/neural-evidence-welfare-code. The code requires only Python, NumPy and Matplotlib. All other material is drawn from the published literature.
Ethics and consent: No human or animal subjects were involved in this research.

# Appendix A. Formal Model

*Supports Sections 3.1 to 3.3 and 5.1.*

**A.1 Decision environment.** Time is discrete. The individual occupies states $s_t \in S$, chooses actions $a_t \in A$, receives reward $r_t$ on the transition out of $s_t$ and moves to $s_{t+1}$ under a kernel $P(s_{t+1} \mid s_t, a_t)$. An intervention $p$ replaces that kernel by $P^p$ and may change the reward and cue structure.

**A.2 Critic.** The critic estimates the value of the decision rule the individual implements, body equation (1), and updates on the prediction error of body equation (3),

$$\theta_{t+1} = \theta_t + \alpha_\theta \, \delta_t \, \nabla_\theta V_\theta(s_t).$$

Equation (9) is a semi-gradient rule and is not the gradient of any objective, because the target is treated as fixed while it depends on $\theta$ (Sutton, 1988). The critic's fixed point is a property of the rule implemented, not the solution to a problem the individual is solving, so a welfare claim cannot be read off it as a revealed objective.

**A.3 Actor.** Actions are drawn from the softmax rule of body equation (2), in which the action-preference function $h_\phi$ ranks actions in reward units. The actor updates on the same prediction error, which is the architecture drawn in Figure 1 (Barto, 1994),

$$\phi_{t+1} = \phi_t + \alpha_\phi \, \delta_t \, \nabla_\phi \log \pi_\phi(a_t \mid s_t), \qquad \nabla_\phi \log \pi_\phi(a_t \mid s_t) = \beta \left[ \nabla_\phi h_\phi(s_t, a_t) - \sum_{a'} \pi_\phi \, (a' \mid s_t) \nabla_\phi h_\phi(s_t, a') \right].$$

The two learning rates are distinct and neither is identified from the other.

**A.4 A normalisation.** The inverse temperature $\beta$ and the scale of $h_\phi$ enter body equation (2) only through their product, so set

$$h_\phi(s, a) = \phi^\top x(s, a), \qquad \|\phi\| = 1,$$

for known features $x(s, a)$, leaving $\beta$ to carry the choice-sensitivity interpretation.

**A.5 Neural encoding.** Body equation (4) gives the dopaminergic channel. The prefrontal channel is

$$n_t^{\text{vmPFC}} = g(V_\theta(s_t)) + \eta_t,$$

and the observation vector is $n_t = (n_t^{\text{DA}}, n_t^{\text{vmPFC}})$.

**A.6 Welfare representation.** Implemented instant utility is the dual-self mixture of body equation (5), the weight $\lambda(s)$ on the long-run self falling in cue states (Fudenberg and Levine, 2006), and the reward that drives learning is that same utility, body equation (6). A criterion $U$ is drawn from an admissible set $\mathcal{U}$, and interventions are scored by (7) and compared by (8). The welfare value of a decision rule $\pi$ at a state, under that criterion, is

$$W_\pi(s;\, U) \;=\; \mathbb{E}^\pi\left[\textstyle\sum_{k=0}^{T} \tilde{\gamma}^k \; U(s_{t+k}, a_{t+k}) \mid s_t = s\right],$$

and the divergence set is

$$D(U) = \left\{ s \in S \colon \pi_\phi(\cdot \mid s) \text{ does not maximise } W_{\pi_\phi}(s;\, U) \right\},$$

which Step 5 asks the analyst to locate. Membership has two sources, a weight below one and a cue distortion away from zero.

**A.7 Cue distortion.** In cue states the action preferences of (11) carry an additional cue-associated term,

$$h_\phi(s,a) \;=\; \phi^\top x(s,a) + \kappa(s)\, C(s,a), \qquad \kappa(s) = 0 \;\text{ for }\; s \notin S^{\text{cue}},$$

with $S^{\text{cue}}$ the set of cue states, $C(s,a)$ the learned cue reactivity of action $a$ and $\kappa(s)$ the strength of the distortion. The distortion reaches the value function through learning, because the critic values the rule implemented and not the rule the individual would endorse.

# Appendix B. Identification

*Supports Sections 2.2, 4.1 and 4.2.*

**B.1 Structures.** A structure is $\sigma = (M, \theta, \phi, \psi)$, with $M$ the model class, $(\theta, \phi)$ the critic and actor parameters and $\psi = (f, g, F_\varepsilon, F_\eta)$ the encoding maps and noise laws. Two structures are observationally equivalent on choice when

$$P(a_{1:T} \mid \sigma) = P(a_{1:T} \mid \sigma') \quad \text{for every action path.}$$

**B.2 What the encoding class buys.** Affine maps with positive slope recover the latent variables up to a common scale and location, and separate two structures whenever they disagree about the magnitude of the predicted signal. Monotone maps recover only ordinal content, and separate two structures only when they disagree about the ranking of the latent variable (Caplin and Dean, 2008). A claim about how much a signal moves is not supported by the weaker assumption.

**B.3 Assumptions.**

- **A1 Measurement noise.** $\varepsilon_t$ and $\eta_t$ are mean zero, independent across channels and over time and of the latent variables and action path, and their law is common to the structures compared with a non-vanishing characteristic function. The law need not be known, since the argument uses only that both structures are convolved with the same injective kernel.

- **A2 Encoding class.** $f$ and $g$ lie in a finite-dimensional class $\Psi$ fixed in advance and shared by the structures compared (B.2).

- **A3 Temporal alignment.** $n_t$ is observed at the resolution of the latent variable it encodes, or the aggregation operator belongs to $\Psi$.

- **A4 Common environment.** The structures are compared on one environment, the observed action path has positive probability under both, and the latent processes are not almost surely constant.

- **A5 Reward and utility link.** Body equation (6).

- **A6 Welfare selection.** $U \in \mathcal{U}$, with $\tilde{\gamma}$ and $T$ set by the evaluator.

**B.4 Proposition 1.** Let $L^a$ and $L^{a,n}$ be the likelihoods of the action path and of the action and neural paths jointly. Suppose (16) holds and let A1 to A4 hold. Then $L^{a,n}(\sigma) \neq L^{a,n}(\sigma')$ if and only if, for some $t$,

$$\mathcal{L}(f(\delta_t),\, g(V_\theta(s_t)) \,|\, a_{1:T},\, \sigma) \;\neq\; \mathcal{L}(f'(\delta_t'),\, g'(V_{\theta'}(s_t)) \,|\, a_{1:T},\, \sigma').$$

*Proof sketch.* Factor $L^{a,n}(\sigma) = P(a_{1:T} \mid \sigma)\, P(n_{1:T} \mid a_{1:T}, \sigma)$. The first factor is common by (16), so the joint likelihoods differ exactly when the second factors do. By A1 the conditional law of $n_{1:T}$ given the action path is a convolution of the conditional law of the encoded latents with a noise law shared by both structures whose characteristic function does not vanish, and such a convolution is injective on distributions (Schennach, 2016). □

*Corollary 1.* If two structures share their encoding maps and induce the same conditional law of the latent path given actions, neural data are uninformative at any sample size.

*Corollary 2, ordinal form.* If the encoding maps are assumed only monotone, a rival is absorbed by refitting the map if and only if it implies the same ordering of the latent variable, since a monotone map preserves order. Monotone encodings therefore separate structures that order it differently and fail only against those that agree on the ordering and differ in the spacing. Separation on magnitudes requires the affine restriction of B.2.

*Remark on power.* Proposition 1 concerns identification. Whether the difference in (17) is detectable at the noise levels and temporal resolution of human neuroimaging belongs to Step 3, and the proposition supplies no power calculation.

**B.5 Remark 1.** Under A5 and A6 the joint likelihood does not depend on which element of $\mathcal{U}$ is selected as $U$, because $U$ enters only body equation (7), which plays no part in generating actions or neural signals. The sign of the comparison in body equation (8) is therefore unidentified whenever $\mathcal{U}$ contains two criteria that rank an intervention oppositely, which any cue-suppressing intervention does once both $u^L$ and $u^S$ are admissible. Identifying the ranking requires a restriction on $\mathcal{U}$ that rests on normative argument.

# Appendix C. Notation

*Supports the whole paper.*

Table C1 lists all 49 symbols in the paper, with their meanings, the equation or section at which each is first defined, and whether each is data, a model parameter, a latent computational variable, a welfare object, a primitive or a statistical object.

Table C1: Notation

| Symbol | Meaning | First defined in | Type |
|---|---|---|---|
| $t$ | Discrete time index | 3.1 | primitive |
| $S$ | State space | 3.1 | primitive |
| $s_t$ | State at $t$ | 3.1 | latent computational variable |
| $A$ | Action set | 3.1 | primitive |
| $a_t$ | Action at $t$ | 3.1 | data |
| $r_t$ | Reward received on the transition from $s_t$ to $s_{t+1}$; $r_t \in \mathbb{R}$ | 3.1 | latent computational variable |
| $P(s_{t+1} \mid s_t, a_t)$ | Transition kernel of the environment | Appendix A | primitive |
| $P^p$ | Transition kernel under intervention $p$ | 3.3, eq. (7) | primitive |
| $\gamma$ | Agent's discount factor | 3.1, eq. (1) | model parameter |

| Symbol | Meaning | First defined in | Type |
|---|---|---|---|
| $V^{\pi_\phi}(s)$ | Value of the implemented decision rule at $s$ | 3.1, eq. (1) | latent computational variable |
| $V_\theta(s)$ | Critic's parametric estimate of $V^{\pi_\phi}$ | 3.1, eq. (1) | latent computational variable |
| $\theta$ | Critic parameters | 3.1, eq. (1) | model parameter |
| $h_\phi(s,a)$ | Actor's action-preference function, in reward units | 3.1, eq. (2) | latent computational variable |
| $\phi$ | Actor parameters | 3.1, eq. (2) | model parameter |
| $\pi_\phi(a \mid s)$ | Policy, that is the decision rule the person implements | 3.1, eq. (2) | latent computational variable |
| $\beta$ | Inverse temperature, sensitivity to differences in action preference | 3.1, eq. (2) | model parameter |
| $\delta_t$ | Temporal-difference error | 3.1, eq. (3) | latent computational variable |
| $\alpha_V$ | Critic learning rate | Appendix A, eq. (9) | model parameter |
| $\alpha_\pi$ | Actor learning rate | Appendix A, eq. (10) | model parameter |
| $n_t$ | Vector of neural measurements at $t$, $(n_t^{\mathrm{DA}}, n_t^{\mathrm{vmPFC}})$ | 3.2 | data |
| $n_t^{\mathrm{DA}}$ | Dopaminergic measurement at $t$ | 3.2, eq. (4) | data |
| $n_t^{\mathrm{vmPFC}}$ | Ventromedial prefrontal measurement at $t$ | Appendix A, eq. (12) | data |
| $f$ | Link function encoding $\delta_t$ into $n_t^{\mathrm{DA}}$ | 3.2, eq. (4) | model parameter |
| $g$ | Link function encoding $V_\theta(s_t)$ into $n_t^{\mathrm{vmPFC}}$ | Appendix A, eq. (12) | model parameter |
| $\varepsilon_t$ | Measurement noise in the dopaminergic channel | 3.2, eq. (4) | data (measurement error) |
| $\eta_t$ | Measurement noise in the vmPFC channel | Appendix A, eq. (12) | data (measurement error) |

| Symbol | Meaning | First defined in | Type |
|---|---|---|---|
| $\psi$ | Encoding parameters, $(f, g, F_\varepsilon, F_\eta)$ | 4.2 | model parameter |
| $\Psi$ | Admissible class of encoding parameters, restricted by A2 | 4.2, A2 | model parameter (class) |
| $U(s,a)$ | Welfare-relevant instant utility, the welfare criterion | 3.3 | welfare object |
| $u^L(s,a)$ | Long-run self's instant utility | 3.3, eq. (5) | welfare object |
| $u^S(s,a)$ | Short-run self's instant utility | 3.3, eq. (5) | welfare object |
| $\lambda(s)$ | Weight on the long-run self in state $s$, in [0,1] | 3.3, eq. (5) | model parameter |
| $U^{\text{impl}}(s,a)$ | Implemented instant utility, the mixture in eq. (8) | 3.3, eq. (5) | latent computational variable |
| $\mathcal{U}$ | Admissible set of welfare criteria | 3.3, A6 | welfare object (class) |
| $W(p;U)$ | Welfare functional for intervention $p$ under criterion $U$ | 3.3, eq. (7) | welfare object |
| $\Delta(p,p_0;U)$ | Welfare comparison of $p$ against baseline $p_0$ | 3.3, eq. (8) | welfare object |
| $\tilde{\gamma}$ | Evaluator's discount factor | 3.3, eq. (7) | welfare object (parameter of the criterion) |
| $T$ | Appraisal horizon | 3.3, eq. (7) | welfare object (parameter of the criterion) |
| $p$ | Policy intervention that changes the decision environment | 3.3, eq. (7) | primitive |
| $p_0$ | Baseline policy | 3.3, eq. (8) | primitive |
| $W_\pi(s;\, U)$ | Welfare value of a decision rule at a state, under criterion $U$ | Appendix A, eq. (13) | welfare object |
| $D(U)$ | Divergence set, states where the implemented rule is not optimal for $U$ | Appendix A, eq. (14) | welfare object |
| $M, M'$ | Two decision and learning model classes under comparison | 4.2 | model parameter (specification) |

| Symbol | Meaning | First defined in | Type |
|---|---|---|---|
| $\sigma, \sigma'$ | Structures, $(M, \theta, \phi, \psi)$ and $(M', \theta', \phi', \psi')$ | 4.2 | model parameter (specification) |
| $L^{a}(\sigma)$ | Likelihood of the action path | Appendix B, eq. (16) | statistical object |
| $L^{a,n}(\sigma)$ | Joint likelihood of the action and neural paths | Appendix B, eq. (16) | statistical object |
| $S^{\text{cue}}$ | Set of cue states | Appendix A, eq. (15) | primitive |
| $C(s, a)$ | Learned cue reactivity of action $a$ in state $s$ | Appendix A, eq. (15) | latent computational variable |
| $\kappa(s)$ | Strength of cue distortion in state $s$ | Appendix A, eq. (15) | model parameter |

# Appendix D. The Checklist, Its Formal Referents and Its Conditions

*Supports Table 1 and Sections 4 and 6.*

Table D1 states, for each step of the checklist, the equations that give it a formal referent, the assumptions it relies on and the condition of Section 4 it operationalises. Two conditions split across two steps each, which is why four yield six.

Table D1

| Step | Formal referent | Assumptions | Condition |
|---|---|---|---|
| 1. Define the welfare criterion $U$ | (7), (8) | A6 | 4.3 Normative transparency |
| 2. Specify the computational model | (1), (2), (3), (5), (6), (9), (10) | A5 | 4.2 Model identifiability |
| 3. Validate neural encodings | (4), (12) | A1, A3 | 4.1 Measurement validity |
| 4. Assess model identifiability | (16), (17), Proposition 1, Corollaries 1 and 2 | A1 to A4 | 4.2 Model identifiability |
| 5. Locate welfare-relevant divergences | (5), (13), (14), (15) | A5 | 4.3 Normative transparency |
| 6. Analyse policy implementation | none | none | 4.4 Policy relevance and institutional context |

Step 6 has no formal referent, which is a feature of the framework and not a gap in it. Whether a classification legitimately restricts a transaction is a question about institutions, and no equation here answers it.

# Appendix E. The Figure 3 Simulation

*Supports Section 3.2.*

Table E1

| **Parameter** | **Value** |
|---|---|
| Discount factor $\gamma$, per within-trial time step | 0.98 |
| Critic learning rate $\alpha_\theta$ | 0.10 |
| Time steps per trial | 20, indexed 0 to 19 |
| Cue onset | step 4 |
| Reward delivery | step 14 |
| Reward magnitude | 1.0 |
| Acquisition trials | 300 |
| Random seed | 20260907 |

Panel B was computed by the author from the model of Section 3.1, using a tapped-delay-line representation of the cue-reward interval and the parameters above. Following body equation (3), the reward is received on the transition out of the current state, so the error a transition generates is recorded at the step the successor state is entered, and the cue response is the error on entering the cue state from a zero-valued baseline. At trial 1 the prediction error is zero at the cue and $+1.0000$ at reward. At trial 300 it is $+0.8171$ at the cue and $0.0000$ at reward, the cue value equalling $\gamma^{10}$. On an omission probe it is $+0.8171$ at the cue and $-1.0000$ at the omitted reward. The code is available at https://github.com/louisyzhu/neural-evidence-welfare-code.

# Appendix F. Sources for the Tobacco Case, by Checklist Step

*Supports Section 5.1.*

Table F1 gives, for each step of the walkthrough, the provision of the Act it rests on, the paragraph or table of the impact assessment, and the page of the command paper. Command-paper pages are printed folios, PDF page in brackets.

Table F1

| Step | Act 2026 (c. 18) | Impact assessment | Command paper CP 949-I |
|---|---|---|---|
| 1. Welfare criterion | not applicable | Option 2 summary sheet; para 213 (disease cases not monetised) | p. 2 [6] |
| 2. Computational model | s. 1(1) | paras 1484 to 1487 (Markov model); para 1485 (England, ages 14 to 30) | p. 20 [24] |
| 3. Validate encodings | not applicable | para 11 fn 22 (Leslie, 2020); para 467 | not addressed |
| 4. Identifiability | not applicable | paras 185, 1515, 1520 (University College London 2021 calibration) | not addressed |
| 5. Locate divergences | s. 1(1); s. 1(2) defence | paras 103, 206, 211; Table 13 | p. 2 [6] |
| 6. Implementation | ss. 16 to 22; s. 165; s. 175(4), (5), (8) | paras 1473 to 1480 (monitoring) | p. 20 [24] |

Rows 3 and 4 carry no command-paper entry and no Act provision, which is the finding the section turns on. Neither document addresses the two steps at which neural evidence could have contributed.